%% file: arxiv.tex
\documentclass[final]{cvpr}

\usepackage{times}
\usepackage{epsfig}
\usepackage{graphicx}
\usepackage{amsmath}
\usepackage{amssymb}
\usepackage{makecell}
\usepackage{multirow}

\graphicspath{{images/}}

\usepackage[pagebackref=true,breaklinks=true,colorlinks,bookmarks=false]{hyperref}

\def\cvprPaperID{5549} 
\def\confYear{CVPR 2021}
\begin{document}

\title{Learning to Sample the Most Useful Training Patches from Images}

\author{
Shuyang Sun$^{1}$ \qquad Liang Chen$^2$ \qquad Gregory Slabaugh$^2$ \qquad
Philip Torr$^1$ \\
$^1$University of Oxford \qquad
$^2$Huawei Noah’s Ark Lab
}

\maketitle
\begin{abstract}
Some image restoration tasks like demosaicing require difficult training samples to learn effective models. 
Existing methods attempt to address this data training problem by manually collecting a new training dataset that contains adequate hard samples, however, there are still hard and simple areas even within one single image. In this paper, we present a data-driven approach called \textbf{PatchNet} that \textbf{learns to select} the most useful patches from an image to construct a new training set instead of manual or random selection. We show that our simple idea automatically selects informative samples out from a large-scale dataset, leading to a surprising \textbf{2.35dB} generalisation gain in terms of PSNR. In addition to its remarkable effectiveness, PatchNet is also resource-friendly as it is applied only during training and therefore does not require any additional computational cost during at inference.
\end{abstract}

\section{Introduction}

Demosaicing and denoising are two indispensable steps of camera image processor pipelines used to produce digital images from RAW sensor measurements. State-of-the-art approaches for joint denoising and demosaicing now rely on deep learning, introduced into the problem by \cite{gharbi2016deepjdd}. However, the performance of deep learning approaches to demosaicing, like that of many other low level computer vision tasks, is highly dependent on the data used to the train the network; and more data does not guarantee better performance. A demosaicing approach trained on one dataset may fail to generalise well to a different dataset.

This large generalisation gap might stem from the difference among distributions of different training datasets. This problem has been studied in \cite{gharbi2016deepjdd}, who perform demosaicing on images of ImageNet \cite{imagenet_cvpr09}, a large-scale natural image dataset. The dataset is dominated by low frequency, smooth patches easy to demosaic, whereas more challenging cases that involve high frequency thin edges and complex textures are less common in the dataset. Therefore most datasets for demosaicing can be described as ``long-tail,'' where the more challenging, and often useful training data is in fact a smaller subset of the complete dataset.  

\input{figures_n_tables/head_pic}
\input{figures_n_tables/fig_method}

One simple solution towards this problem is to manually gather a training dataset that contains enough hard samples. However, such a solution has three major problems. First, collection of a large-scale dataset is both time and labor intensive. Second, even though the whole image may be hard to demosaic in terms of some metrics like PSNR, the image may still contain simple low frequency regions, which might also bring great randomness when training with randomly cropped patches from each image. Third, if the dataset consists only of hard samples then the performance of easier samples will drop. To manually control the difficulty of the training set is a much harder problem since it could be hard to measure the difficulty itself. All these challenges point to an intuitive question, that is, is it possible to construct a training dataset by automatically selecting appropriate samples from a larger dataset?

Selection of a more informative sample for training
to enhance the modeling ability of a learning algorithm is a well-formulated problem called Active Learning (AL). Recently, progress within the field of data-driven active learning \cite{sener2017active, sener2018active, woodward2017active} has already shown the effectiveness of AL in many related problems.
In this paper, instead of proposing a hand-crafted method to resolve the long-tail problem, we introduce a data-driven approach based on active learning to automatically select the most suitable patches from an image to train. As shown in Fig. \ref{fig:head_pic}, each patch used for training will be assigned with a weight measured by PatchNet, such that those patches which are not helpful will be down-graded and ignored during training. 
In addition to PatchNet, we also propose RestoreNet, a new network architecture that applies the structural knowledge extracted from PatchNet. Trained with PatchNet, RestoreNet is able to achieve the state-of-the-art performance on joint denoising and demosaicing.

Overall, the contributions of this paper are three-fold:

\noindent (1) We propose PatchNet, which effectively resolves the long-tail problem by automatically learning the non-negative \textit{trainability}
of each training sample. Notably, the proposed method has no additional cost during testing.

\noindent (2) We achieve state-of-the-art performance for the tasks like demosaicing and joint demosaicing and denoising with the application of PatchNet. We also show that with the help of PatchNet, we can boost the generalisation performance by a large margin (\textbf{+2.35dB} in terms of PSNR) when the distributions of the training and testing sets differ.

\noindent (3) We propose RestoreNet, a new network architecture that achieves the state-of-the-art performance even without the help of PatchNet.

\section{Related Work}
\label{sec:relate}

\noindent \textbf{Joint denoising and demosaicing}. The joint denoising and demosaicing (JDD) task is a fundamental problem in the image signal processor pipeline in digital photography.  The JDD task seeks to compute missing color values for noisy data collected on a color filter array (CFA) such as the Bayer pattern.  The problem has been addressed using traditional spatial and frequency domain methods \cite{li2008demosaicing}, as well as energy minimization \cite{klatzer2016learning} techniques. Gharbi et al. \cite{gharbi2016deepjdd} proposed a deep, data-driven framework called DemosaicNet that relies on a CNN to solve this problem. Demosaicing is an interesting problem in that many patterns in RGB images are lower frequency ``easy'' cases which may only require simple processing such as bilinear interpolation. ``Hard'' examples, such as high frequency patterns occur less frequently, and this leads to a training data imbalance, so learning-based models are likely to struggle in these hard examples. DemosiacNet provides a large dataset consisting of hard examples for JDD was developed comprehensively. This dataset is commonly used as a benchmark for demosaicing and JDD. This paper also proposed a novel CNN architecture for JDD leveraging the characteristics of Bayer patterns of raw images. Most recently, a self-guidance network (SGNet) \cite{sgnet} was proposed to extend DemosiacNet. Advanced components, e.g., residual-in-residual dense blocks were used to enhance the CNN capability. A dense image map was proposed to use to guide the CNN on hard patterns (e.g. high frequency lines). This could be limited in cases that objects have weak edges.

In addition to data-driven methods, model-driven methods have been explored as well. For instance, adaptive filters in the CNNs could be learned to handle different image patterns \cite{kokkinos2019aiterative}, \cite{kokkinos2019biterative}. More specifically, a dictionary of filters and a CNN are iteratively learned so that the resulting network performs well on different types of patterns. The limitation here is at the inference stage, as the filters should be selected and then used, which could be expensive in terms of runtime and/or memory.

Instead of learning the JDD task on single images, raw image bursts \cite{wronski2019handheld, ehret2019joint} can be leveraged.  \cite{wronski2019handheld} combines multiple raw frames with subpixel shifts, whereas \cite{ehret2019joint} presents a self-supervised technique where supervision in training comes from small motions between raw images in a burst. In both \cite{wronski2019handheld} and \cite{ehret2019joint}, accurate motion estimation between frames is vital.

\input{figures_n_tables/hist}
\noindent\textbf{Active learning for computer vision.}
The goal of Active Learning (AL) is to learn to judge the informativeness of each training sample and pick out those  considered to be useful for training.
Recent active learning methods rely on an acquisition function that estimates the sample informativeness with a learned metric \cite{konyushkova2017learning, wang2017cost}. Meanwhile, reinforcement learning has gained attention as a method to learn a labelling policy that directly maximizes the learning algorithm performance \cite{mnih2013playing, liu2018learning, bachman2017learning, pang2018meta, padmakumar2018learning, contardo2017meta, sener2018active}.

\noindent However, all these methods are designed for high-level vision tasks in which manual labels are given. To our knowledge, active learning for low-level vision tasks has rarely been explored compared to other computer vision tasks.
 The reason may lie in that acquiring ground truth in low level may require considerable human effort, thereby most low-level vision datasets do not provide human labels, which makes it hard to bridge the AL methods from the high-level to the low-level. In this paper, we transform the problem into a metric learning problem.
\input{figures_n_tables/fig_res_block}

\noindent\textbf{Network architectures.}
The success of AlexNet \cite{alexnet} on the ILSVRC challenge \cite{imagenet} catalyzed a surge of interest in deep learning. However, sequentially connected networks like \cite{vggnet} and \cite{googlenet} soon met the bottleneck of depth. In order to solve this problem, He et al. \cite{resnet, resnext} introduced residual learning that sums up the features within the same resolution into deep networks to enable networks to be extremely deep. An alternative approach to solve this problem is to densely concatenate the layers in the same stage \cite{densenet}.
As a downstream task, network architectures for low-level vision basically follow the corresponding trend in high-level vision, \eg, sequentially stacked convolutions \cite{srcnn, espcn, asrcnn}, residual learning \cite{dncnn, edsr, mdsr, drrn, carn}, dense connectivity \cite{srdensenet, rdn, dbpn}. A more comprehensive review about the recent trend of deep network architectures for low-level vision tasks is summarized in \cite{srsurvey}.

\section{PatchNet}
\label{sec:method}
\subsection{Manually Mining Hard Patches from Images}
Before introducing our method, we first demonstrate the long-tail problem within the demosaicing problem. 
As shown in Figure \ref{fig:hist}, most patches restored by DeepJDD \cite{gharbi2016deepjdd} are of high quality (PSNR $> 40$).
To find the hard samples for training, we first attempted  hard-negative mining manually by finetuning the pre-trained network \cite{gharbi2016deepjdd} with different thresholds. In this way, samples with PSNR lower than the threshold (hard samples) are kept during training, while those with higher PSNR are skipped.
%
Though the manual hard negative mining could solve this problem to an extent, there are two major problems remaining:

(1) During training, the quality of the restored image is changing due to the optimization of the network parameters. This may lead to great fluctuation of the mining threshold. To manually adjust the threshold per iteration is hard and ungeneralizable to different data-dependent filters.

(2) Some hard samples solely selected by PSNR have low contextual information, while the trainability is a high-level metric that needs to be measured with contextual information. If the dataset is awash with these samples, then the training dataset could be misleading.

Therefore, to the end of automatically and filter-adaptively distinguishing the hard samples from the entire dataset, we propose a network named PatchNet learning such process.
The PatchNet is used to reweight patches within the image, correspondingly, the backbone network that is responsible for restoring the original image is called as RestoreNet in this paper.

\subsection{Network architecture}
As shown in Figure \ref{fig:method}, we feed the output (usually a RGB image) of the RestoreNet into the PatchNet as the input. By down-sampling each patch with a pre-defined size into a single pixel that represents the trainability of that patch, we formulate the problem as a supervised regression problem.

PatchNet consists of multiple stages that gradually transform the image into a set of trainability scalars, one for each patch of the original image. Each stage is composed of several convolutional blocks and a down-sampling layer. The number of stages $N_{stage}$ is calculated by:
\begin{equation}
    N_{stage} = \log_{2} k,
\end{equation}
where $k$ denotes the width (in pixels) of the square patch. Here we use the popular residual block \cite{resnet} that consists of two $1\times1$ convolutions and another $3\times3$ convolution as the basic building block displacing the pure convolution for feature extraction. Detailed design of the residual block is presented in Figure \ref{fig:res_block}. Note that unlike most low-level vision pipelines that remove all batch normalizations \cite{ioffe2015batchnorm} from the model, we keep them because  PatchNet is modeling a 
high-level pattern, e.g. the trainability. We modify the traditional activation function from ReLU into leaky ReLU with a slope factor $\alpha=0.2$ when the value is negative.
We denote each patch at location $p$ on the output restored image $I$ from the RestoreNet as $I_p$, for which the trainability scalar is denoted as $\mathbf{t}_p$, then the whole process could be described as:
\begin{equation}
    \mathbf{t}_p = \phi(f(I_p; \mathbf{W}_f)),
\end{equation}
where $f$ and $\mathbf{W}_f$ represent PatchNet and its corresponding learnable weights within the whole network, and $\phi$ represents the binarization function. Normally $p$ on the input image should be a square-shaped bounding box with size $k\times k$.
To imitate the selection operation, the trainability scalars should be binarized. Here we use a sigmoid function $\phi(\cdot)$ with temperature $T$ controlling the sharpness of the function for binarization, which could be formulated as:
\begin{equation}
    \phi(\mathbf{x}) = \frac{1}{1+e^{-T\mathbf{x}}},
    \label{eq:func_phi}
\end{equation}
where $\mathbf{x}$ denotes the input.
The binarized output of the function $\phi(\cdot)$ will be also serve as the final output of PatchNet. In this way, we obtain an individual trainability value $t_p$ for each input patch $p$.

\subsection{The loss function}
Intuitively, to guide the RestoreNet with the trainability value, we could apply $\mathbf{t}_p$ back onto the output of the RestoreNet $I_p$ during the loss computation.
Assume that the original loss function for the RestoreNet of each patch is $\mathcal{L}^R$, and the total loss for both the PatchNet and the RestoreNet as $\mathcal{L}_p$. The loss could be calculated by:
\begin{equation}
    \mathcal{L}_p = \frac{N}{\sum_N^p \mathbf{t}_p}\mathbf{t}_p\mathcal{L}^R_p,
\end{equation}
where the constant number $N$ represents the number of patches we crop from every single image. The loss weight $\frac{N}{\sum_N^p \mathbf{t}_p}$ here rescales the loss to its original magnitude. In this case, the original loss will be dispatched onto patches with larger trainability. The loss that penalizes the RestoreNet could be diverse. In this paper, we choose L2 loss as the loss function for RestoreNet, and compute the corresponding final loss based on that.

\input{figures_n_tables/table_patchnet}

\subsection{RestoreNet}
In principle, PatchNet can be used in conjunction with any data-dependent image restoration network. In this paper, we exploit the knowledge we learnt from constructing PatchNet, and use its building block (without BN) demonstrated in Figure \ref{fig:res_block} as the basic block of the network for restoration. During implementation, the bottleneck ratio $r$ is set to be $2$, and the number of channels $C$ is set to be 256. The depth (number of basic blocks) is 16. The entire network is constructed by replacing all convolutions in DeepJDD \cite{gharbi2016deepjdd} into the proposed building blocks.

\section{PatchNet on JDD training}
Given a clean RGB image $I_{linRGB}$ in linear space (without nonlinear operators, e.g. gamma correction), its associated noisy raw image is usually formulated as
\begin{equation}
    I_{nRaw} = M \circ I_{linRGB} + \mathcal{N}(\sigma),
\end{equation}
Here, $M$ is the mask represnting the camera Bayer pattern, \eg RGGB, which degrades $I_{linRGB}$ into the raw domain. $\mathcal{N}(\cdot)$ is the pixel independent Gaussian noise parameterized by the variance $\sigma$. Without the noise, $I_{cRaw} = M \circ I_{linRGB}$ is the clean image in the raw domain. To this end, the $I_{nRaw}$ is the noisy raw image output from the sensor. JDD task learns a mapping from $I_{nRaw}$ to $I_{linRGB}$, restoring the linear RGB images from the sensor raw, conditioned on the $\sigma$:
\begin{equation}
    I_{linRGB} = J(I_{nRaw};\theta|\sigma).
\end{equation}

The $J(\cdot)$ was modelled using a CNN as per \cite{gharbi2016deepjdd} which could serve as a RestoreNet as mentioned before. The predicted linear RGB image $\hat{I}_{linRGB}$ is compared against the ground truth image on pixel intensities, for instance,
\begin{equation}
    \mathcal{L}^R = || \hat{I}_{linRGB} - I_{linRGB} || _2.
\end{equation}
In this case, each training image (patch) is treated equally and might not lead to satisfactory results.

Based on the conventional $\mathcal{L}^R$, the proposed PatchNet learns a quasi-binary map, so only useful patches are actually involved in the training process.

\section{Experiments}
\label{sec:experiment}
\input{figures_n_tables/table1}
\input{figures_n_tables/table2}
\subsection{Datasets}
Due to our real-world interest in generalization, we need to collect two datasets containing natural images and only hard samples respectively. For the natural images, we choose the large-scale dataset Vimeo-90K \cite{vimeo} as the benchmark. The dataset originally is built for evaluating various video restoration tasks. With 89,800 independent clips from 4,278 videos with resolution of $448 \times 256$, the dataset is both large and diverse enough for us to select as the natural image dataset. We select the $4th$ image  from each clip which has 7 frames in total. The whole dataset is split into 64,612 clips for training and 7,824 clips for evaluation.

For the dataset with only hard samples, we choose the MIT Moire \cite{gharbi2016deepjdd}. The dataset is carefully collected from a extremely large parent set downloaded from the web. By applying a network trained on ImageNet to millions of new patches, failure cases are detected and retained to compose the MIT Moire dataset. There are 2.6 million patches with size of $128\times 128$ within the dataset.

For the task of JDD, to further validate the effectiveness of our approach, we add two other datasets with high resolution images for training and evaluation respectively. The DIV2k dataset \cite{div2k}, which contains 800 2K resolution images, is selected as part of the training set. Meanwhile the Urban100 dataset \cite{urban100} with 100 high resolution images is chosen as one of our test sets.

\subsection{Implementation details}
\input{figures_n_tables/table_patchsize} \input{figures_n_tables/table_temperature}
\textbf{Technical Details for PatchNet.} As shown in Table \ref{tab:table_patchnet}, we construct two different networks that differ in numbers of convolutional layers (depths). We name them as PatchNet-tiny and PatchNet-large according to their corresponding numbers of convolutions.
As described in Section \ref{sec:method}, the input of PatchNet is a full-size image with padding. The size of each patch $k \times k$ within the image is $64\times 64$, and the sum of both the horizontal and vertical padding is set to be $64$. Note that the padding size for each side of the image is a random number less than $64$ for each iteration. The random padding could be regarded as a way of data augmentation to generate more patches with different patterns. Note that the PatchNet is only applied during training, so there is no need to apply all these tricks in testing phase. The network could be also trained with randomly cropped patches. In this case, there is no need to add random paddings for getting the randomness.

\textbf{Technical details for training.} We optimize the network with an Adam optimizer \cite{adam}. The learning rate is initialized to be $2.5\times 10^{-4}$ and is adjusted per epoch with a half-cycle cosine schedule. The batch size is set to be $16$ for all experiments. We train the network for 50 epochs for pure demosaicing and 150 epochs for JDD.  As for Joint Denoising \& Demosaicing, the input is perturbed with noise $\sigma \in [0, 16]$. As the datasets we used for training vary in size (800 images for DIV2K, 89,800 images for Vimeo), we over-sample the DIV2K dataset by 20 times to resolve the imbalance.
All experiments are conducted on a single GPU (NVIDIA Tesla V100).

\subsection{Ablation Studies}
\input{figures_n_tables/table3}
We conduct our ablation studies primarily on demosaicing and then extend them to JDD. In this section, all experiments are conducted for demosaicing if not specified.

\textbf{Verifying the Long-Tail Problem with Manual Hard Negative Mining.} We first validate our motivation through manual hard-negative sampling. As introduced in Section \ref{sec:method}, we filter out those hard sample through thresholding the values of patches in terms of PSNR. We tried different thresholds from $35$ to $45$ that only retain samples below the threshold. The network is first pre-trained with all the data within the Vimeo-90k for 50 epochs, and then finetuned with only the hard samples for another 100 epochs.

Shown in Table \ref{tab:table1}, the manual hard negative mining could lead to $0.29$dB improvement. We also compare results that training the network on the natural image dataset Vimeo-90k and the hard dataset MIT Moire training set separately while testing both models on the same MIT Moire testing set, but got extremely different results ($31.38$dB vs. $37.0$dB).
All these results indicate that the long-tail problem we assume truly exists. Thereby, a solution towards this problem is needed.

\textbf{Evaluating PatchNet on Demosaicing.} Table \ref{tab:table2} lists the results achieved by directly applying PatchNet onto the baseline when training the network on the natural image dataset Vimeo-90k~\cite{vimeo}. 
The comparison between the baseline and the baseline with PatchNet indicates that our PatchNet leads to large improvement (\textbf{2.35dB}) in terms of generalization to hard samples (from MIT Moire). PatchNet is not only effective for such domain adaptation settings.  It can also boost training performance when trained and evaluated within the same domain. This is validated by the experiments on Vimeo-90k. A clear difference (\textbf{0.89dB}) is shown between the settings of baseline with and without PatchNet.


\textbf{The effect of patch size.} Patch size is an important hyper-parameter that  significantly affects the performance of the entire framework. Table \ref{tab:patch_size} exhibits that the performance improves when the patch size gets larger. However, PatchNet with patch size 128 will lead to higher  cost. We thereby choose 64 as the final patch size of PatchNet.

\input{figures_n_tables/fig_patches}
\textbf{The effect of Temperature $T$ in function $\phi(\cdot)$.} The temperature we set in Equation \ref{eq:func_phi} controls the sharpness of the binarization function $\phi(\cdot)$. A higher temperature produces a sharper curve of the function $\phi(\cdot)$. Table \ref{tab:temperature} shows the effects for different temperatures for the task of pure demosaicing on Vimeo-90k. Based on these experimental results, we conclude that the binarization function cannot be either too sharp or too flat.
So we choose 2 as the final setting of the temperature.

\input{figures_n_tables/fig_sigma5}
\input{figures_n_tables/fig_sigma15}
\textbf{Evaluating PatchNet on JDD.} 
Experimental results for PatchNet in the task of Joint Denoising and Demosaicing (JDD) is reported in Table \ref{tab:table3}. PatchNet together with the RestoreNet designed with our own knowledge is able to surpass all state-of-the-art methods on three challenging datasets, \eg Vimeo-90k, MIT Moire and DIV-2k. Here we emphasize that the PatchNet has no cost at test time since it is only proposed during training. 
 Shown in Table \ref{tab:table3}, we could find that PatchNet performs better for patches corrupted by more severe noise, which indicates that our PatchNet is particularly helpful for more challenging cases when $\sigma$ becomes larger. This validates our assumption on PatchNet that it might be useful for resolving the long-tail problem.
Unlike demosaicing, the gain that PatchNet achieves on JDD is less pronounced. This is not hard to understand. For demosaicing, most of the patches collected from the natural images could be already well solved by the data-dependent filter, which makes it hard for us to harvest the knowledge from the hard samples.  This long-tail distribution is less significant in the JDD problem. For most patches with noise and mosaics, their PSNR values are already low enough for the network to define them as hard samples.
This indicates that even without PatchNet, random selection from images is already capable to find hard patches for the JDD problem to train.

\subsection{Visualization on Patches Weighted by PatchNet}
To better verify our motivation and check what kind of patterns the PatchNet actually learns, we visualize the learnt output of PatchNet in Figure \ref{fig:patches}. From Figure \ref{fig:patches}, we could observe that:

(1) The outputs of PatchNet are changing during the training of RestoreNet, which indicates that PatchNet is highly adaptive to the capacity of the backbone model. This is reasonable since the capability of RestoreNet is increasing during training, and many previously identified hard samples will be gradually become easy ones. This is demonstrated in Figure \ref{fig:patches} where the number of blue patches for epoch 35 (the second row) is fewer than that for epoch 15 (the first row).

(2) Patches with high-frequency content are more likely to be kept by PatchNet, whereas those with low-frequency content tend to be ignored by PatchNet during training. This phenomenon also makes sense because highly contrastive contents, \eg edges, grids, stripes, are more likely to generate visual artifacts, rendering  these regions harder to restore.

\subsection{Visualized Comparison Results}
We further visualize our results and compare them with the state-of-the-art. Figures \ref{fig:viz_jdd_sigma5} and \ref{fig:viz_jdd_sigma15} demonstrate the effectiveness of our proposed RestoreNet and PatchNet. With the inclusion of PatchNet, the baseline backbone is able to focus on the hard patches during training, and therefore achieves better results, restoring with more detail areas with high-frequency patterns like edges, rasters, and grids.

\section{Conclusion}
In this paper, we introduce PatchNet, which is used to resolve the long-tail problem encountered in some image restoration tasks like demosaicing and JDD. By adaptively reweighing the loss magnitude with an estimated trainability scalar for each patch during training, our PatchNet is able to steadily alleviate the long-tail problem. This leads to a surprising improvement (over $2$dB in terms of PSNR) when the gap between the training and testing sets is large. Apart from PatchNet, we also construct RestoreNet, a new backbone network utlizing structural knowledge identified by PatchNet. We validate our backbone network and also PatchNet on three challenging datasets for image restoration tasks \eg demosaicing, denoising and JDD. Experimental results show that our model can surpass all state-of-the-art baselines without any extra computational cost during testing.

\input{acknowledgment}

{\small
\bibliographystyle{ieee_fullname}
\bibliography{references}
}
\end{document}

%% file: figures_n_tables/head_pic.tex
\begin{figure}
    \centering
    \includegraphics[scale=0.666]{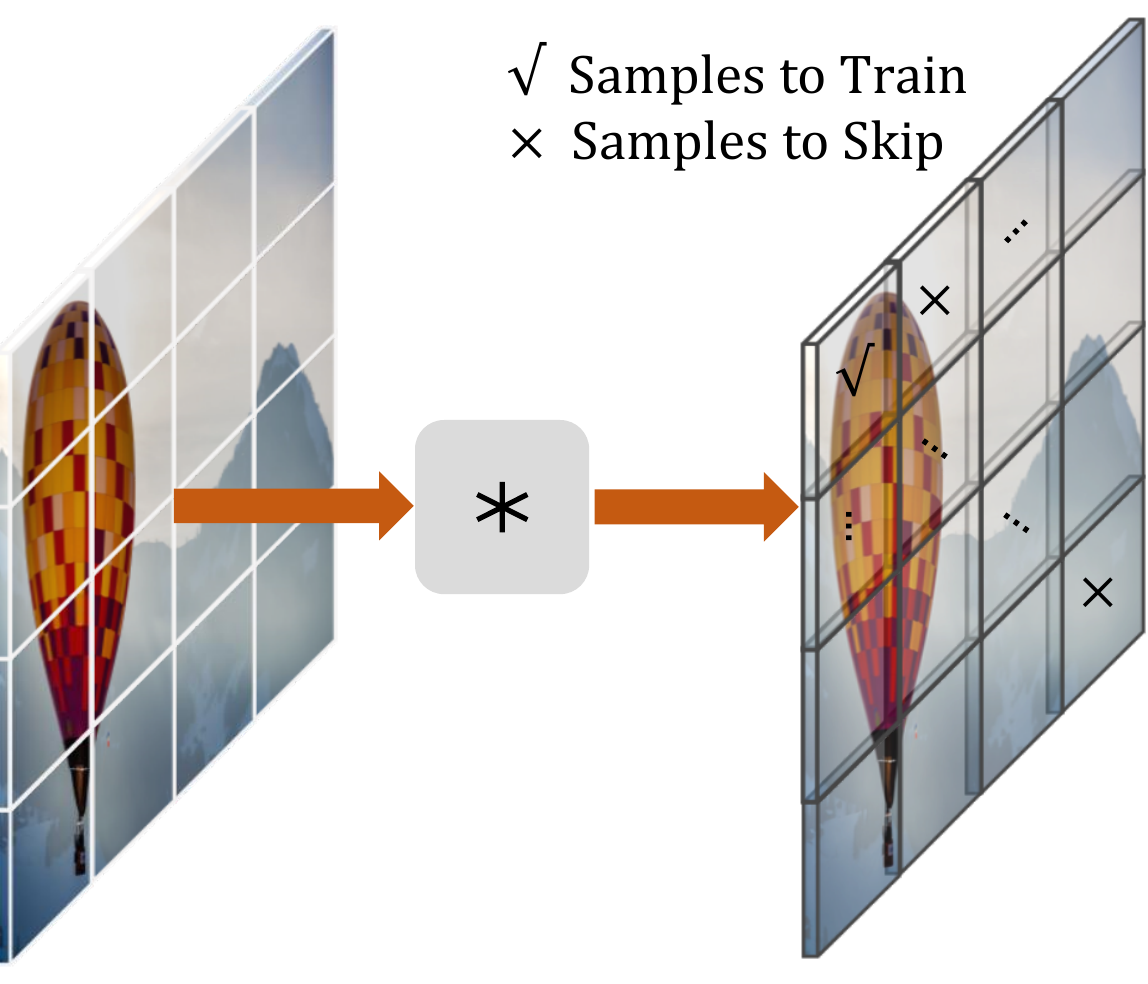}
    \caption{A broad overview of the proposed method. The proposed method learns different trainability weights for different image parts or patches during training. The trainability weights are encoded on a per-patch basis, forming a trainability map on the right-hand side. The map will be then used as evidence guiding the learning process of the image restoration.}
    \label{fig:head_pic}
\end{figure}

%% file: figures_n_tables/fig_method.tex
\begin{figure*}
    \centering
    \includegraphics[scale=0.666]{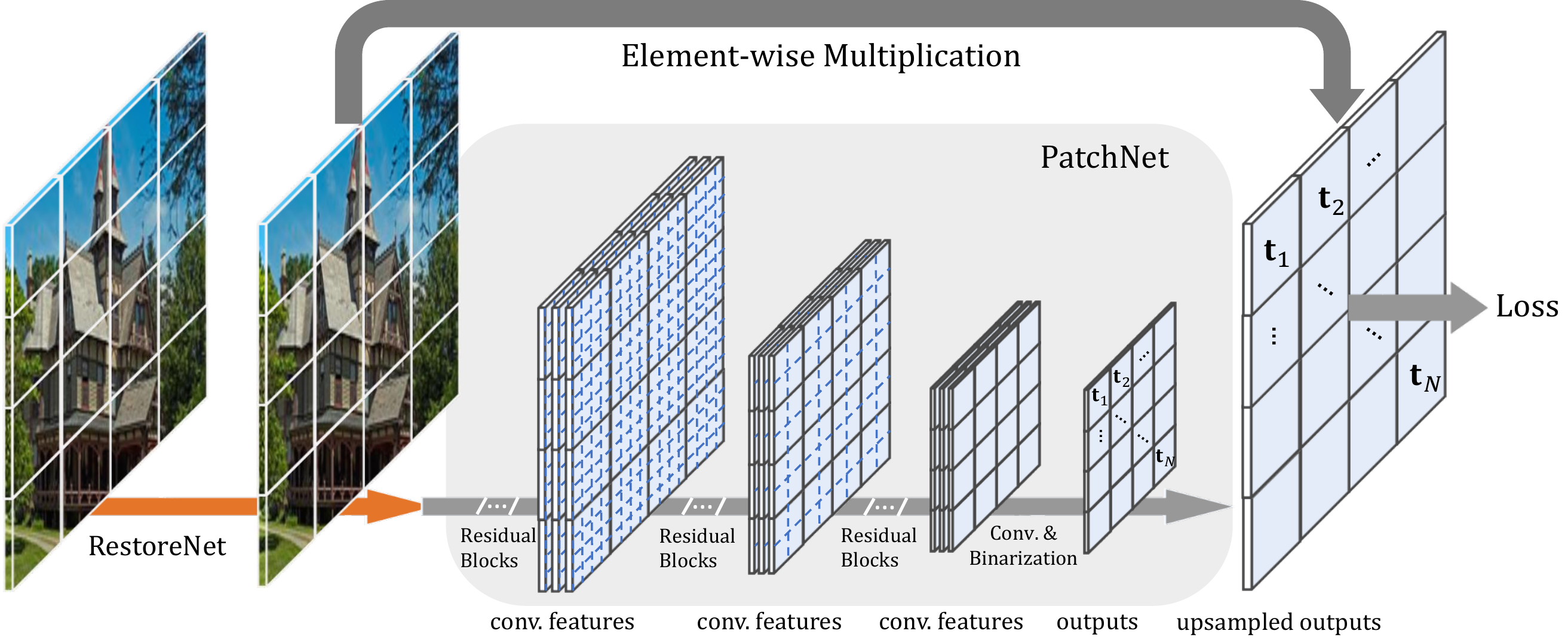}
    \caption{The architecture of the proposed method. PatchNet is a feed-forward network with a series of stages. Each stage contains several convolutions and a down-sampling operator, so each input patch will be finally down-sampled to one single trainability scalar. The trainability scalar will be binararized and applied back onto the entire patch through an attention-like element-wise multiplication operation. The masked output will be then supervised by a loss defined in Section \ref{sec:method}.  The whole framework is trained end-to-end.}
    \label{fig:method}
\end{figure*}

%% file: figures_n_tables/hist.tex
\begin{figure}
    \centering
    \includegraphics[scale=0.5]{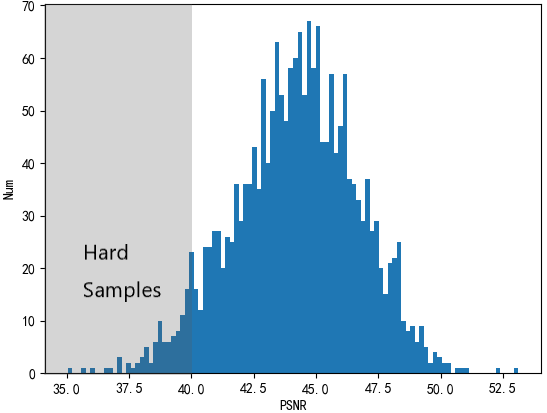}
    \caption{Histogram for 2000 patches restored by \cite{gharbi2016deepjdd} from the Vimeo-90k validation set. Most patches are of very high quality (PSNR $>$ 40).}
    \label{fig:hist}
\end{figure}

%% file: figures_n_tables/fig_res_block.tex
\begin{figure}
    \centering
    \includegraphics[scale=0.888]{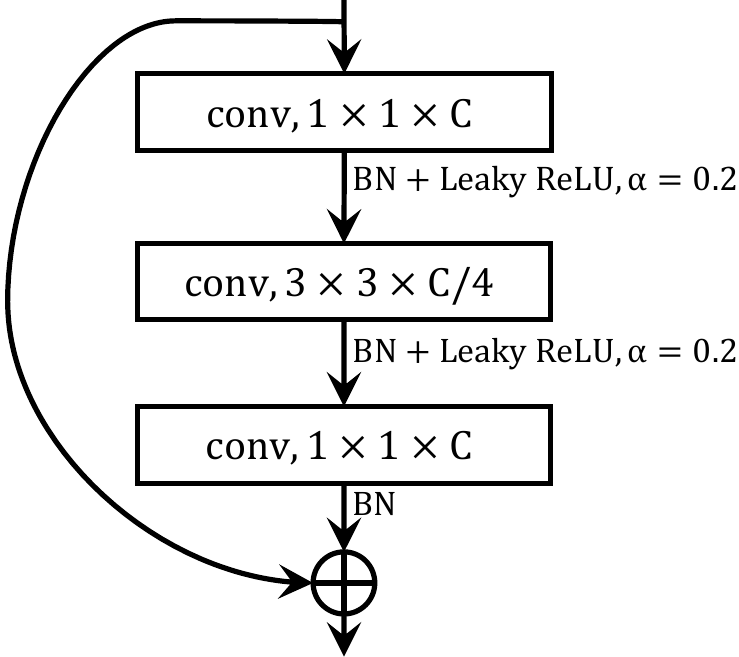}
    \caption{Our revised residual block for PatchNet and RestoreNet. The denotation $3\times 3 \times C$ indicates the layer is convolved with a $3\times 3$ kernel and the number of output channels is $C$. We choose a $4 \times$ bottleneck structure on the channel dimension to save computational cost. Notably, we replace all ReLU activations with leaky ReLU parameterized by a negative slope $\alpha=0.2$, and remove the activation function for the last convolution.}
    \label{fig:res_block}
\end{figure}

%% file: figures_n_tables/table_patchnet.tex
\begin{table}[t]
\setlength{\tabcolsep}{1.6pt}
    \centering
    \scalebox{0.88}{
    \begin{tabular}{c|c|c|c}
    \Xhline{1.5pt}
        Layer & Output Size & PatchNet-Tiny & PatchNet-Large\\
        \hline
        Input & $H\times W\times 3$ & \multicolumn{2}{c}{$1\times1, 64$, Conv}\\
        \hline
        $Stage_1$ & $\frac{H}{2}\times \frac{W}{2}\times 64$ 
        & \makecell[c]{
        $\begin{bmatrix} 1\times1, 16 \\ 3\times3, 16 \\ 3\times3, 64 \end{bmatrix} \times 1$ \\ Avg Pool} & \makecell[c]{$\begin{bmatrix} 1\times1, 16 \\ 3\times3, 16 \\ 3\times3, 64 \end{bmatrix} \times 3$ \\ Avg Pool}\\
        \hline
        $Stage_2$ & $\frac{H}{2}\times \frac{W}{2}\times 64$ 
        & \makecell[c]{
        $\begin{bmatrix} 1\times1, 16 \\ 3\times3, 16 \\ 3\times3, 64 \end{bmatrix} \times 1$ \\ Avg Pool} & \makecell[c]{$\begin{bmatrix} 1\times1, 16 \\ 3\times3, 16 \\ 3\times3, 64 \end{bmatrix} \times 3$ \\ Avg Pool}\\
        \hline
        $Stage_3$ & $\frac{H}{8}\times \frac{W}{8}\times 128$ 
        & \makecell[c]{
        $\begin{bmatrix} 1\times1, 32 \\ 3\times3, 32 \\ 3\times3, 128 \end{bmatrix} \times 1$ \\ Avg Pool} & \makecell[c]{$\begin{bmatrix} 1\times1, 32 \\ 3\times3, 32 \\ 3\times3, 128 \end{bmatrix} \times 4$ \\ Avg Pool}\\
        \hline
        $Stage_4$ & $\frac{H}{16}\times \frac{W}{16}\times 64$ 
        & \makecell[c]{
        $\begin{bmatrix} 1\times1, 64 \\ 3\times3, 64 \\ 3\times3, 256 \end{bmatrix} \times 2$ \\ Avg Pool} & \makecell[c]{$\begin{bmatrix} 1\times1, 64 \\ 3\times3, 64 \\ 3\times3, 256 \end{bmatrix} \times 6$ \\ Avg Pool}\\
        \hline
        $Stage_5$ & $\frac{H}{32}\times \frac{W}{32}\times 64 $
        & \makecell[c]{
        $\begin{bmatrix} 1\times1, 128 \\ 3\times3, 128 \\ 3\times3, 512 \end{bmatrix} \times 2$ \\ Avg Pool} & \makecell[c]{$\begin{bmatrix} 1\times1, 128 \\ 3\times3, 128 \\ 3\times3, 512 \end{bmatrix} \times 6$ \\ Avg Pool}\\
        \hline
        $Stage_6$ & $\frac{H}{64}\times \frac{W}{64}\times 64$ 
        & \makecell[c]{
        $\begin{bmatrix} 1\times1, 256 \\ 3\times3, 256 \\ 3\times3, 1024 \end{bmatrix} \times 1$ \\ Avg Pool} & \makecell[c]{$\begin{bmatrix} 1\times1, 256 \\ 3\times3, 256 \\ 3\times3, 1024 \end{bmatrix} \times 3$ \\ Avg Pool}\\
        \hline
        Upsample & $H\times W\times 3$ & \multicolumn{2}{c}{\makecell[c]{$1\times 1, 3$, Conv \\ Binarization \& Upsampling}}\\
        \Xhline{1.5pt}
    \end{tabular}}
    \caption{Detailed architecture for PatchNet. Components within the bracket represent the configurations of convolutions (kernel size and number of output channels) of the Res-Block. Detailed structure of the building block with BN and ReLU is illustrated in Figure \ref{fig:res_block}. Here $H, W$ represent the height and width of the input image, and the number of stages $6$ is equivalent to $log_{2}64$, where $64$ is the patch size.}
    \label{tab:table_patchnet}
\end{table}

%% file: figures_n_tables/table1.tex
\begin{table}[t]
    \centering
    \begin{tabular}{ccc}
    \Xhline{1.5pt}
        Method & Threshold & \makecell[c]{Output \\PSNR (dB)}\\
        \hline
        Baseline \cite{gharbi2016deepjdd} & - & 44.20 \\
        \textit{w/} HNM & 35.0  & 44.27 \\
        \textit{w/} HNM & 40.0  & \textbf{44.49} \\
        \textit{w/} HNM & 42.5  & 44.41 \\
        \textit{w/} HNM & 45.0  & 44.41 \\
        \textit{w/} HNM & 45-35  & 44.48 \\
        \Xhline{1.5pt}
    \end{tabular}
    \caption{PSNR values when applying Hard-Negative Mining (HNM) with different thresholds on Vimeo-90k. The threshold for the last row is linearly scaled per epoch from 45 to 35 (dB).}
    \label{tab:table1}
\end{table}

%% file: figures_n_tables/table2.tex
\begin{table}[t]
\setlength{\tabcolsep}{5.6pt}
    \centering
    \begin{tabular}{cccc}
    \Xhline{1.5pt}
        Method & \makecell[c]{Vimeo-90k \\ (dB)} & \makecell[c]{MIT Moire \\ (dB)}\\
        \hline
        Kokkinos et al. \cite{kokkinos2018deep} & 44.37 & 31.44 & \\
        \hline
        DeepJDD \cite{gharbi2016deepjdd} & 44.20 & 31.38 \\
        \textit{w/} HNM & 44.49 & 31.87 \\
        \textit{w/} PatchNet-tiny & \textbf{44.90} (+0.70) & \textbf{33.01} (+1.63) \\
        \textit{w/} PatchNet-large & \textbf{45.09} (+0.89) & \textbf{33.73} (+2.35) \\
        \Xhline{1.5pt}
    \end{tabular}
    \caption{Evaluation results on MIT Moire and Vimeo-90k testing set when methods are trained on Vimeo-90k training set.}
    \label{tab:table2}
\end{table}

%% file: figures_n_tables/table_patchsize.tex
\begin{table}[t]
    \centering
    \begin{tabular}{ccc}
    \Xhline{1.5pt}
        Method & Patch Size & \makecell[c]{Output \\PSNR (dB)}\\
        \hline
        Baseline \cite{gharbi2016deepjdd} & - & 44.20 \\
        \textit{w/} PatchNet & 16  & 43.07 \\
        \textit{w/} PatchNet & 32  & 44.68 \\
        \textit{w/} PatchNet & 64  & 45.09 \\
        \textit{w/} PatchNet & 128  & 45.11 \\
        \Xhline{1.5pt}
    \end{tabular}
    \caption{PSNR values for PatchNet-large with different patch sizes on Vimeo-90k. Note that when the patch size is 128, we will add another stage (with convs \& weights) in PatchNet for down-sampling. For other PatchNet with smaller patch size (16 and 32), we just remove the average pooling layer in existing stages.}
    \label{tab:patch_size}
\end{table}

%% file: figures_n_tables/table_temperature.tex
\begin{table}[t]
    \centering
    \begin{tabular}{ccc}
    \Xhline{1.5pt}
        Method & $T$ & \makecell[c]{Output \\PSNR (dB)}\\
        \hline
        Baseline \cite{gharbi2016deepjdd} & - & 44.20 \\
        \textit{w/} PatchNet & 1 & 44.65 \\
        \textit{w/} PatchNet & 2 & \textbf{45.09} \\
        \textit{w/} PatchNet & 5 & 45.01 \\
        \textit{w/} PatchNet & 10 & 44.93 \\
        \Xhline{1.5pt}
    \end{tabular}
    \caption{PSNR values for PatchNet with different $T$ of function $\phi(\cdot)$ defined in Equation \ref{eq:func_phi}.}
    \label{tab:temperature}
\end{table}

%% file: figures_n_tables/table3.tex
\begin{table}[t]
\setlength{\tabcolsep}{6.6pt}
    \centering
    \scalebox{0.88}{
    \begin{tabular}{l|c|c|c|c}
    \Xhline{1.5pt}
        Method & $\sigma$ & \makecell[c]{Vimeo-90k\\(dB)} & \makecell[c]{MIT Moire\\(dB)} & \makecell[c]{Urban100 \\(dB)}\\
        \hline
        Kokkinos \cite{kokkinos2018deep} & \multirow{6}{*}{5} & - & 31.94  & 34.07 \\
        SGNet \cite{sgnet} & & - & 32.15  & 34.54 \\
        CDM* \cite{cdm} & & - &  30.36 & 32.09 \\
        \cline{1-1} \cline{3-5}
        DeepJDD \cite{gharbi2016deepjdd} & & 39.61  & 31.82  & 34.04 \\
        \makecell[r]{\textit{w/} PatchNet} & & 39.73  & 31.91  & 34.02  \\
        \cline{1-1} \cline{3-5}
        Ours  & & 39.75 & 32.33  & 34.62  \\
        \makecell[r]{\textit{w/} PatchNet}  & & \textbf{39.80} & \textbf{32.38}  & \textbf{34.66}  \\
        \Xhline{1.0pt}
        Kokkinos \cite{kokkinos2018deep} & \multirow{6}{*}{10} & - & 30.01  & 31.73 \\
        SGNet \cite{sgnet} & & - & 30.09  & 32.14 \\
        CDM* \cite{cdm} & & - &  28.63 & 30.03 \\
        \cline{1-1} \cline{3-5}
        DeepJDD \cite{gharbi2016deepjdd} & & 36.36  & 29.75  & 31.60 \\
        \makecell[r]{\textit{w/} PatchNet} & & 36.49  & 29.88  & 31.74 \\
        \cline{1-1} \cline{3-5}
        Ours  & & 36.61  & 30.26 & 32.20 \\
        \makecell[r]{\textit{w/} PatchNet}  & & \textbf{36.71}  & \textbf{30.31}  & \textbf{32.33} \\
        \Xhline{1.0pt}
        Kokkinos \cite{kokkinos2018deep} & \multirow{6}{*}{15} & - & 28.28  & 29.87 \\
        SGNet \cite{sgnet} & & - & 28.60  & 30.37 \\
        CDM* \cite{cdm} & & - &  27.23 & 28.34 \\
        \cline{1-1} \cline{3-5}
        DeepJDD \cite{gharbi2016deepjdd} & & 34.44  & 28.22  & 29.73 \\
        \makecell[r]{\textit{w/} PatchNet} & & 34.67  & 28.38  & 29.99 \\
        \cline{1-1} \cline{3-5}
        Ours  & & 34.66 & 28.74  & 30.42  \\
        \makecell[r]{\textit{w/} PatchNet}  & & \textbf{34.93} & \textbf{28.87}  & \textbf{30.58} \\
        \Xhline{1.5pt}
    \end{tabular}
    }
    \caption{Evaluation results for JDD on three challenging datasets. Our method is able to achieve the state-of-the-art performance. Note that the PatchNet here will not be used for testing, which means it is totally cost-free for all models trained with PatchNet.}
    \label{tab:table3}
\end{table}

%% file: figures_n_tables/fig_patches.tex
\begin{figure}
\setlength{\tabcolsep}{3.6pt}
    \centering
        \begin{tabular}{ccc}
        \includegraphics[scale=0.28]{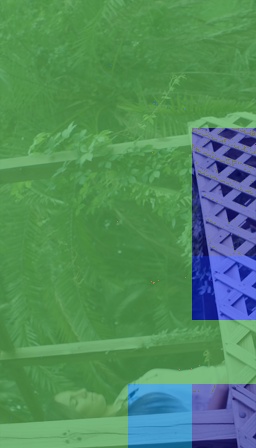} & \includegraphics[scale=0.28]{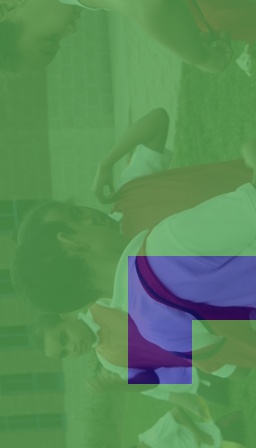} & 
        \includegraphics[scale=0.28]{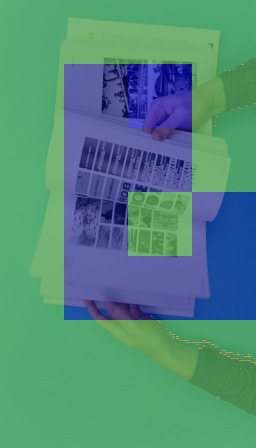} \\
        \includegraphics[scale=0.28]{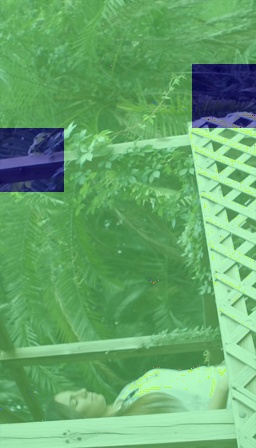} & \includegraphics[scale=0.28]{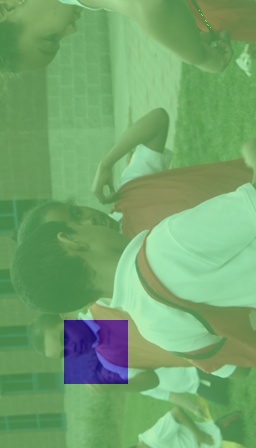} & 
        \includegraphics[scale=0.28]{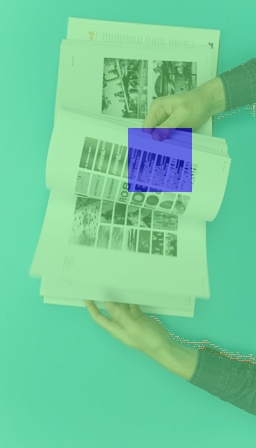}
        \end{tabular}
    \caption{Visualization of the learnt output of PatchNet. Patches rendered with a blue overlay are used in training, however patches rendered in green are tend to be ignored.
    The first row: visualized outputs for epoch 15. The second row: visualized outputs for epoch 35. We could see that the highlighted patches are generally with more high-frequency patterns.}
    \label{fig:patches}
\end{figure}

%% file: figures_n_tables/fig_sigma5.tex
\begin{figure}
    \centering
    \includegraphics{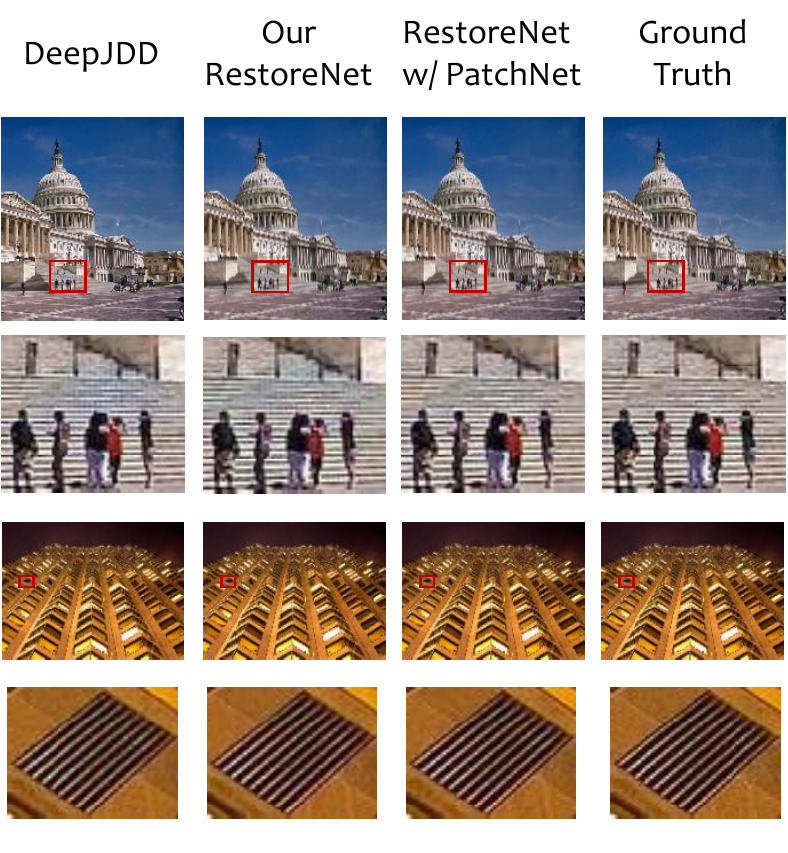}
    \caption{Visualized comparison for JDD when $\sigma=5$.}
    \label{fig:viz_jdd_sigma5}
\end{figure}

%% file: figures_n_tables/fig_sigma15.tex
\begin{figure*}[t]
    \centering
    \includegraphics[scale=0.78]{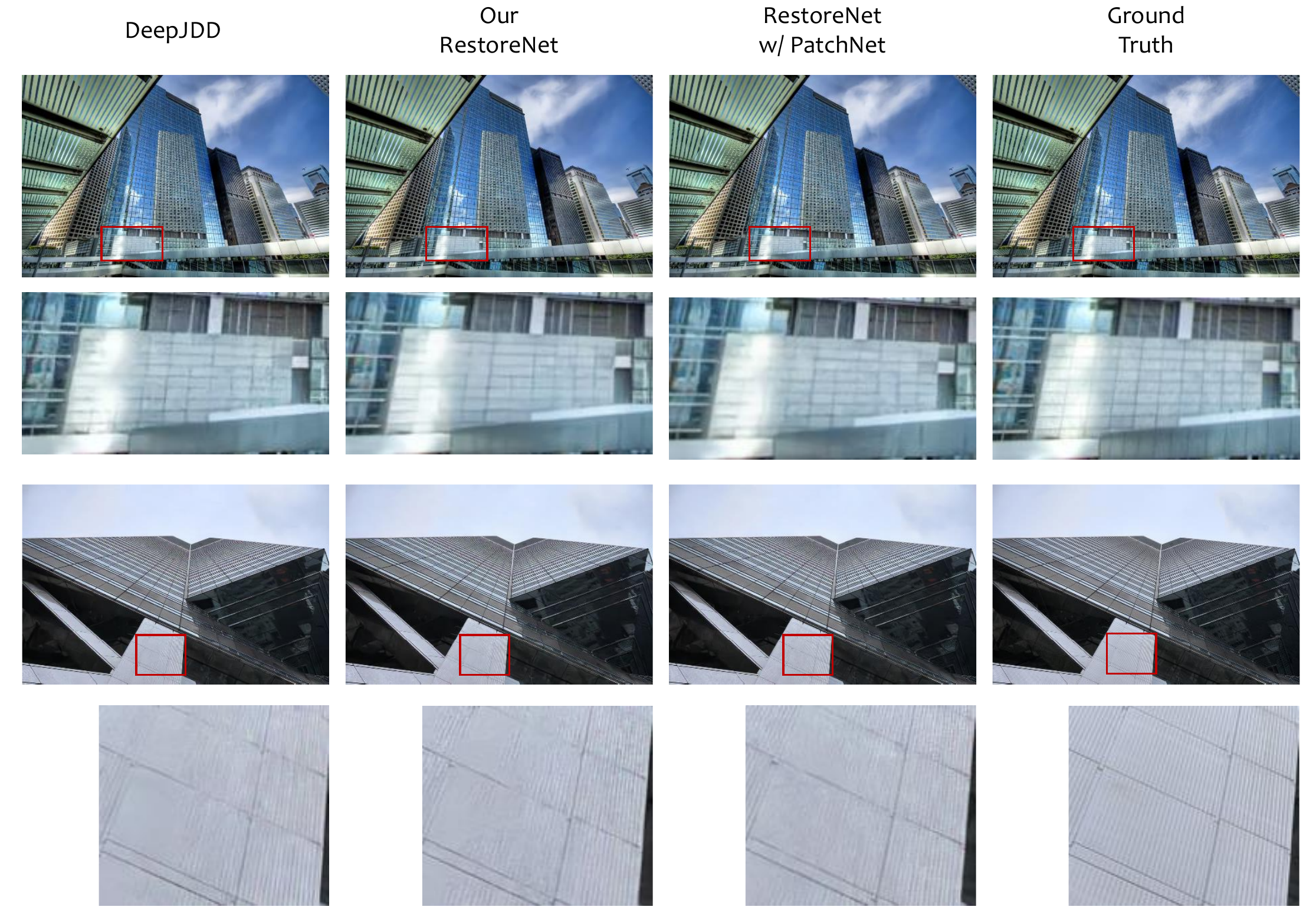}
    \caption{Visualized comparison for JDD when $\sigma=15$. Best viewed in color and zoomed in.}
    \label{fig:viz_jdd_sigma15}
\end{figure*}

%% file: acknowledgment.tex
\noindent \textbf{Acknowledgements} We would like to thank Andrew Gambardella and Yifu Tao for proof-reading and helpful comments. This work is supported by Huawei Technologies Co., Ltd., the ERC grant ERC-2012-AdG
321162-HELIOS, EPSRC grant Seebibyte EP/M013774/1 and EPSRC/MURI grant EP/N019474/1. We would also like to thank the Royal Academy of Engineering and FiveAI.